# UTILIZING XAI TECHNIQUE TO IMPROVE AUTOENCODER BASED MODEL FOR COMPUTER NETWORK ANOMALY DETECTION WITH SHAPLEY ADDITIVE EXPLANATION(SHAP)

Khushnaseeb Roshan and Aasim Zafar

Department of Computer Science, Aligarh Muslim University,
Aligarh-202002, India

*ABSTRACT*

*Machine learning (ML) and Deep Learning (DL) methods are being adopted rapidly, especially in computer network security, such as fraud detection, network anomaly detection, intrusion detection, and much more. However, the lack of transparency of ML and DL based models is a major obstacle to their implementation and criticized due to its black-box nature, even with such tremendous results. Explainable Artificial Intelligence (XAI) is a promising area that can improve the trustworthiness of these models by giving explanations and interpreting its output. If the internal working of the ML and DL based models is understandable, then it can further help to improve its performance. The objective of this paper is to show that how XAI can be used to interpret the results of the DL model, the autoencoder in this case. And, based on the interpretation, we improved its performance for computer network anomaly detection. The kernel SHAP method, which is based on the shapley values, is used as a novel feature selection technique. This method is used to identify only those features that are actually causing the anomalous behaviour of the set of attack/anomaly instances. Later, these feature sets are used to train and validate the autoencoderbut on benign data only. Finally, the built SHAP_Model outperformed the other two models proposed based on the feature selection method. This whole experiment is conducted on the subset of the latest CICIDS2017 network dataset. The overall accuracy and AUC of SHAP_Model is 94% and 0.969, respectively.*

## 1. INTRODUCTION

The rapid increase of digitization, internet traffic, online data transfer and much more made cyberspace vulnerable to unknown attacks. Hence anomaly-based detection systems become essential tools to detect these unknown cyber-attacks effectively [1]. Anomalies in the network are the unusual network traffic behaviour that does not have known signatures in the attack detection system. Network anomalies can arise due to any reason such as network attack, weakness within the system, internal or external network misconfiguration and much more. The problem of anomaly detection has been the focus of the research community since the last two decay [2]–[5]. According to a recent survey [6], many researchers are working on ML and DL techniques to build anomaly-based detection systems because these methods can handle complex data such as network traffic data. But the questions arise when we can not understand the decision making process or prediction of the DL based model (especially in unsupervised learning) due to its opaque nature (it is like a black box to us).

The problem of explainability (or interpretability) of ML and DL is not new. It had existed since the 1970s when researchers were working to explain the output of expert systems [7]. However, the term explainable AI (XAI) was introduced in 2004 by Van Lent [8] in games application and simulation. Initially, this issue was treated seriously, but later it slowed down, and the focus shifted towards improving the accuracy of the models and developing new algorithms. Recently, explainable AI has been again an increase in interest among practitioners and researchers [9], especially for DL based architecture such as autoencoder (AE) and other complex models. The SHAP framework [10] based on shapley values is one of the XAI techniques used in this paper to explain and improve the results of the autoencoder model for network anomaly detection.

Autoencoders are widely used unsupervised neural network architectures for anomaly detection [11][12][13]. In general, the autoencoder is trained on normal/benign data only. Consequently, they can reconstruct benign data with less reconstruction error, but for attack data, it gives a large reconstruction error and provides a major deviation from the benign data. The reconstruction error of any instance can explain anomalies but up to some extent only. One may suspect that features having large reconstruction errors cause this anomalous prediction. But the large error on one feature $(x_i)$ may stem from an anomalous behaviour of another feature, $(x_j)$ [11]. Just by looking only at the raw reconstruction error of each feature, we cannot find the cause of the anomaly. Hence, shapley values would be useful to detect and explain the anomalies as it provides the true contribution of each feature in model prediction [14]. The features having large shapley values are the cause of large reconstruction error and, consequently, become more important than other features for anomaly detection model [14].

The contribution of the paper is as follows:

- A novel approach of feature selection method is proposed based on the Shapley Additive Explanation (SHAP), one of the XAI techniques. The kernelExplainer method is used to build the explanation model for the actual autoencoder model.
- Based on the kerenelExplainer method subset of features are identified that are causing large reconstruction errors of the attack instances. And these features are further used to build an improved version of the network anomaly detection model (SHAP_Model).
- We illustrated how the XAI technique could be used to explain the black-box model output and improve its performance to some extent, as we did in this paper in network anomaly detection.

## 2. RELATED STUDY

This section elaborates on the recent studies proposed for model explanation and its interpretability on anomaly detection.

Amarasinghe et al. [15] proposed a framework based on deep neural networks (DNNs) to answer the following questions "why an instance is anomalous?", "what is the certainty?" and "what are the relevant factors making this prediction?". This explanation further reduces the opaque nature of the deep learning model and ease its adaptability in a real-world environment. One of the ways of detecting and explaining anomalies are outliers. Outliers are also rare as anomalies. And, the same approach can be applied in anomaly explanation as well.

Micenková et al. [16] proposed a method to explain the outlier based on subspaces or attribute subsets. This subset of features provides additional information on detected outliers. The authors used the outlier scoring function to approximate subspaces and a fast heuristic search approach. Dang et al. [17] used the discriminative features approach to detect and interpret outliers in high dimensional data. A proposed method is based on graph embedding and modelling the geometrical structure of data by creating the neighbourhood graph. Similarly, Liu et al. [18] proposed a framework, Contextual Outlier INterpretation (COIN), to interpret the outliers. The authors used three-factor, i.e., outlier score, features that contributed to the outlier, and neighbourhood description to interpret outliers. The authors applied this approach on variousdatasets, synthetic and real datasets such as WBC, Twitter and MNIST. However, the same approach can be extended incorporating heterogeneous data sources, using hierarchical clustering, and applying it in deep learning. Tang et al. [19] focused on improving the interpretability of outliers incorporating other contextual information such as a set of attributes contributing to its unusual behaviour, outlier degree, refereeing group and outlier groups.

Goodall et al. [20] proposed situ to detect anomalous behaviour in streaming network traffic with data visualization. The proposed system is scalable and can provide contextual information that explains the abnormal behaviour of network traffic and logs. Moreover, Collaris et al. [21] presented a case study for fraud detection using random forest. The authors provided the instance level explanation of fraud using the dashboard. Pang and Aggarwal [22] addressed the issue related to the unbounded and lack of supervisory nature of the anomaly. Despite giving such tremendous results, DL-based models are often criticized due to their lack of interpretability. Alvarez-Melis and Jaakkola [23] proposed the self-explaining model with built-in interpretability in its architecture. This approach is different from the posterior model explanation, where the idea is to build a simpler model that learns the local input and output behaviour of the actual model.

To the best of our knowledge, we found the three most relevant studies that explained unsupervised anomalies. Out of three, two are based on autoencoders [24][25], and others explain anomalies using PCA[14]. However, none of the relevant studies implements the idea of using shapley values to improve autoencoder performance in an anomaly detection system.

## 3. BACKGROUND

### 3.1. Autoencoder

Autoencoder is an artificial neural network (ANN) unsupervised architecture, first proposed by Rumelhart et al. [26]. Autoencoders are considered as self-supervised architecture since the inputs and the reconstructed outputs are the same. It consists mainly of two ANN architectures, i.e. Encoder and Decoder. The Encoder encodes the data, and the Decoder decodes the same data with minimum reconstruction error [27]. And, the typical nonlinear activation functions used between the hidden layers are ReLU and Sigmoid [27]. Furthermore, as defined by Goodfellow et al. [28], an autoencoder is "a neural network that is trained to attempt to copy its input to its output". Fig. 1 shows the general diagram of autoencoder architecture with the bottleneck layer. The number of hidden layers and neurons in each layer may differ depending upon the different problem scenarios.

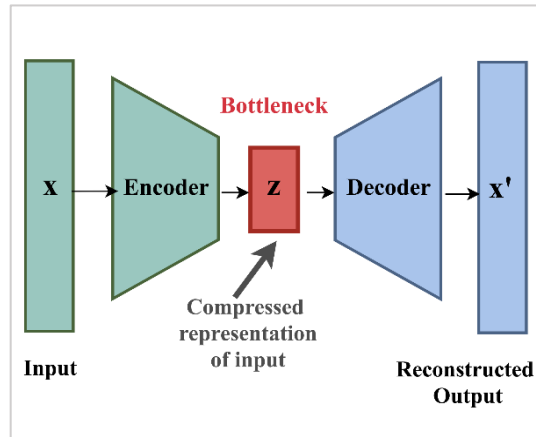

Figure 1. Autoencoder achitecure

As illustrated in Equation (1), Encoder function $\Phi$ maps the given input data to its latent representation $F$ and Decoder function $\Psi$ maps the latent space $F$ to the reconstructed output (recreates the input).

$$\Phi : X \to F$$
$$\Psi : F \to X$$
$$\Phi, \Psi = argmin \, || X - (\Phi \, o \, \Psi) \, ||^2 \quad (1)$$

In general, the reconstruction error of an autoencoder is the difference between the given input $x$ and the generated output $\bar{x}$. And the most common functions to calculate reconstruction error are Mean Squared Error (MSE) and Mean Absolute Error (MAE), as illustrated in Equation (2) and (3).

$$MSE = \sum_{i=1}^{N} (\bar{x} - x)^2 \quad (2)$$

$$MAE = \sum_{i=1}^{N} |\bar{x} - x| \quad (3)$$

Initially, autoencoders were developed for data compression and dimensionality reduction techniques, but nowadays, it is used in many areas such as denoising [29], anomaly detection [30][31], word semantics [32] and much more.

### 3.2. Explainable Artificial Intelligent (XAI)

Explainable artificial intelligence aims to develop a set of tools, techniques, and strategies to produce more transparent, accountable, and explainable models while retaining its powerful predictions [33]. And, in making life-changing decisions such as disease diagnosis, it's crucial to understand why the system makes such a critical decision. Hence the importance of explaining the AI system becomes clear at this point [34]. Furthermore, the black-box nature of the AI-based system gives excellent results but without any explanation, and hence, they lose their trust to adapt these systems in critical decision making [34].

Similarly, the unbounded and lack of supervisory nature of network anomaly sometimes becomea major obstacle to adopt DL- based model, and it becomes difficult to find the cause of anomaly in

International Journal of Computer Networks & Communications (IJCNC) Vol.13, No.6, November 2021

unsupervised learning. Several XAI methods are illustrated in [34], and the SHAP framework is one of the methods discussed in the next section.

### 3.3. Shapley Additive Explanation (SHAP)

SHapley Additive exPlanation (SHAP) is the unified approach for model interpretation proposed by Lundberg and Lee [10]. It is a model agnostic approach proposed to address the interpretability of complex models such as ensemble methods and deep neural networks. SHAP framework combines the previously proposed techniques such as LIME [35] and DeepLIFT [36] under the additive feature attribution class, and methods belonging to this class have explanation models with a linear function of binary variables.

Let $f$ is the actual model, and to explain it, we need to define another simpler model, $g$ (explanation model). To explain the single instance $x$, $g$ uses its simplified version $x'$ with mapping function $h$ such that $x = h(x')$. The explanation model function is defined as in (4).

$$g(z') = \phi_0 + \sum_{i=1}^{M} \phi_i z_i' \qquad (4)$$

Here, $g(z')$ is the explanation model, $z'$ is the simplified input such that $(z') \approx (x')$ and $z' \in \{0,1\}^M$ $\phi_i \in R$ [10].

Furthermore, the SHAP framework uses the concept of shapley values [37], a method from coalitional game theory, to explain the contribution of each feature in model output. Lundberg [10] explained three desirable properties of classical shapley value estimation methods: local accuracy, missingness, and consistency. It is a challenging task to compute the exact shapley values, but the author approximates these values under the additive feature attribution methods. And to calculate the effect of each feature on model prediction, the function is defined as in (5).

$$\phi_i = \sum_{S \subseteq F\{i\}} \frac{|S|! F(|F| - S - 1)!}{|F|!} \left[ f_{S \cup \{i\}}(x_{S \cup (i)}) - f_S(x_S) \right] \qquad (5)$$

Here, $F$ represents all feature sets, and $S$ is a subset of $F$. $f_{S \cup \{i\}}$ is the model trained with $S$ and $i^{th}$ feature, $f_S$ is trained model without this feature and $x_S$ represents the values of features in the set $S$. The difference $f_{S \cup \{i\}}(x_{S \cup (i)}) - f_S(x_S)$ is calculated on all possible subset $S \subseteq F\setminus\{i\}$ and finally, the computed shapley values are the weighted averages of all possible differences. And this representation is used as feature attributions.

In our approach, the kernel SHAP, a model agnostic method is used, to compute the contribution of each feature in the autoencoder reconstruction error for a single instance and overall model on the sample data. The complete procedure is described in section 5.

### 3.4. Evaluation Metrics

The evaluation of the methods has been performed with the following metrics. For any ML / DL based classifier, the results can be classified into four groups [38].

- True Positive (TP): refers to the samples correctly identified as positives.
- True Negative (TN): refers to the samples correctly identified as negatives.

113

International Journal of Computer Networks & Communications (IJCNC) Vol.13, No.6, November 2021

- False Positive (FP): refers to the samples incorrectly identified as positives.
- False Negative (FN): refers to the samples incorrectly identified as negatives.

The following metrics are based on the above ones and can be calculated as:

1) Accuracy (ACC): refers to the ratio between correctly predicted samples to all predicted samples.

$$ACC = \frac{TP + TN}{TP + TN + FP + FN} \qquad (6)$$

2) Recall (Sensitivity or True Positive Rate): refers to the ratio between true positive samples to the actual positive samples.

$$Recall\ (R) = \frac{TP}{TP + FN} \qquad (7)$$

3) Precision (P): refers to the ratio between true positive samples to all positive predicted samples.

$$Precision\ (P) = \frac{TP}{TP + FP} \qquad (8)$$

4) F-Score: it is the harmonic mean of both precision and recall.

$$F - Score = \frac{2 \times R \times P}{R + P} \qquad (9)$$

5) False Positive Rate (FPR): refers to the ratio between false-positive samples to the actual negative samples.

$$False\ Positive\ Rate = \frac{FP}{FP + TN} \qquad (10)$$

6) Specificity (True Negative Rate): refers to the ratio between true negative samples to the actual negative samples.

$$Specificity = \frac{TN}{TN + FP} \qquad (11)$$

7) Receiver Operating Characteristics (ROC): The ROC curve [39] is a standard metric, especially for imbalance class dataset evaluation [40]. The ROC curve provides an efficient summary of the classifier on the range of TPRs and FPRs. It is also used to decide the best threshold for an optimal result. In our case, G-mean [41] [42] is used to select the best threshold for the classification of all three proposed models. G-mean is the square root of Recall and Specificity and is treated as an unbiased classification metric with a chosen threshold.

$$G - mean = \sqrt{Recall \times Specificity} \qquad (12)$$

Further, the area under the ROC curve (AUC) is also used for evaluation as it provides the single value that helps in comparing different models in general. The AUC value lies between 0 and 1; however, it indicates an unrealistic classification if the value is less than 0.5 [43]. Further, as a rough guide for classification, it can be understood as follows: Excellent for 0.9 to 1, Good for 0.80 to 0.90, fair for 0.70 to 0.80, poor for 0.60 to 0.70, and fail for 0.50 to 0.60 AUC.

## 4. DATASET

In order to evaluate the model, the subset of the latest CICIDS2017 dataset is used [44]. The Canadian Institute for Cybersecurity (CIC) produced this dataset within five working days from Monday to Friday in a network emulated environment. It contains a wide range of real-time attacks both in packet-based and flow-based format. We prefered this dataset over KDDCUP99 and NSLKDD due to the following reasons: 1) it is the latest and up to date dataset with a variety of attacks that have recently been carried out on the network [45], 2) It is labelled and contain both flow-based and packet-based format, 3) It contains real-time network traffic characteristics [45]. KDDCUP99 is a highly redundant and very old dataset. The redundancy issue was resolved in NSLKDD. Still, it does not represent realistic network traffic data [46].

Table 1. CICIDS2017 dataset

| Day | Class Type |
|---|---|
| Monday | Benign |
| Tuesday | Benign, FTP-Patator, SSH-Patator |
| Wednesday | Benign, DoS slowloris, DoS Slowhttptest, DoS Hulk, DoS GoldenEye, Heartbleed |
| Thursday | Benign, Web Attack-Brute Force, Web Attack- SQL Injection, Web Attack-XSS |
| Friday | Benign, Bot, DDoS, PortScan |

The CICIDS2017 dataset contains fourteen attack classes: FTP-Patator, SSH-Patator, DoS slowloris, DoS-Slowhttptest, DoS-Hulk DoS-GoldenEye, Heartbleed, Brute-force, XSS, SQL Injection, Infiltration, Bot, DDoS and Port Scan. A detailed analysis of this dataset is available in [47][48]. The only drawback is the imbalanced class distribution of attack samples. However, we have used the subset of the CICIDS2017 dataset initially for our experiment. Table 1 shows a brief description of the CICIDS2017 dataset. In our future study, we will further cover the complete dataset for evaluation with the same proposed approach.

## 5. PROPOSED APPROACH, ALGORITHM AND MODELS

### 5.1. Dataset Preprocessing

The CICIDS2017 files have been downloaded, and a subset of the complete dataset is used for this experiment, as already discussed. Normal/benign data is used for model training without the target class label, and both normal and attack data are used for testing purposes. Table 2 shows the classwise count of data after sampling. Monday file contains only benign network traffic data and is used to train and validate the model. Friday file includes both benign and attack samples and is used as testing data. Total 150000 instances are used to train and validate, and 137718 instances for testing the model. Benign data are split as 67% for training and 33% for validation sets, as shown in Fig. 2. And finally, the StandardScaler function is used to scale all the data.

Table 2. CICIDS2017 dataset after sampling

| Category | Class Label | Instances |
|---|---|---|
| Training Data (Monday) | BENIGN | 150000 |
| Testing Data (Friday) | BENIGN | 97718 |
| | DDoS | 40000 |
| Total | | 287718 |

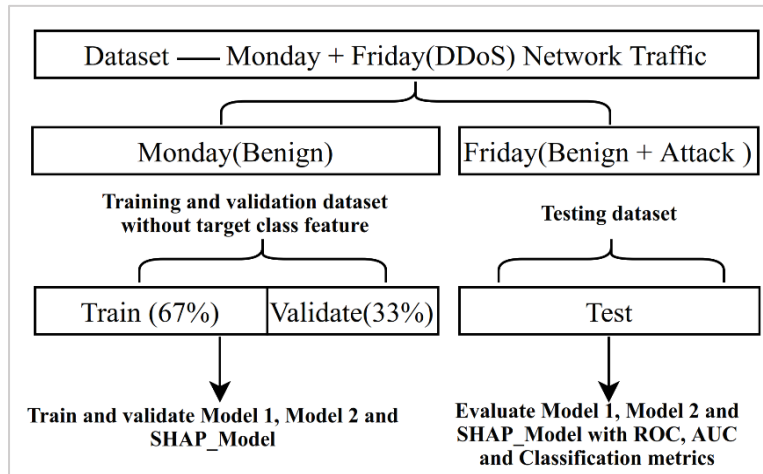

Figure 2. Dataset splitting procedure

Total three approaches are used to build three separate models, namely Model_1, Model_2 and SHAP_Model. The proposed three methods are as follows:

### 5.1.1. First Approach

In the first case, all the features of the training dataset are used for Model_1 training and testing. The correlation between the features is also shown in Fig. 3. Correlation is the dependence or predictability between two variables.

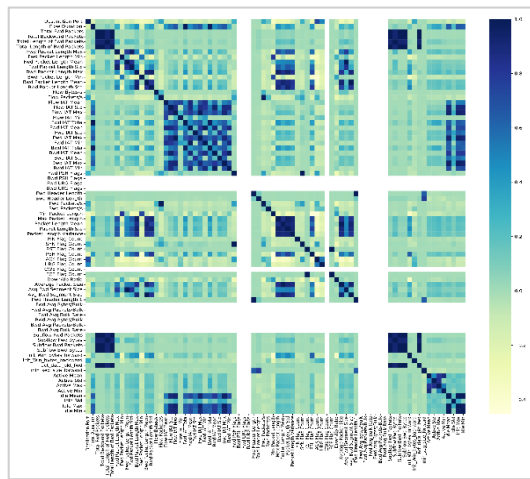

Figure 3. Correlation matrix of all features in the training dataset

The correlation coefficient is used to measure the positive or negative correlation between two variables, and its value lies between -1.0 and +1.0. The dark colour indicates the high correlation, and the light colour indicates the low correlation among the features [49].

### 5.1.2. Second Approach

In the second case, the unsupervised feature correlation approach is used that removed redundant features to improve the model performance and stability [49] [50]. All the features with a correlation greater than 0.8 have been dropped from the dataset, as shown in Algorithm 1 and Fig. 4. And based on the selected 39 features, the Model_2 is built.

**Algorithm 1:** Unsupervised feature selection with correlation approach for Model 2

*Input*: benign_dataset, N
*Output* : selected_feature_data
cols ← benign_dataset.cor().shape[0]
selected_features←{}
**for** i in cols
    **for** j in range (i+1, cor.shape[0])  i,j $\in N$  i$\leq$j
    **if** cor.iloc[i,j] >= 0.8
    **if** cols[j]:
                cols[j] = False
    **end if**
    **end if**
        **end for**
**end for**
selected_features = benign_dataset.columns[cols]
selected_feature_data=benign_dataset[selected_features]
**return** selected_feature_data

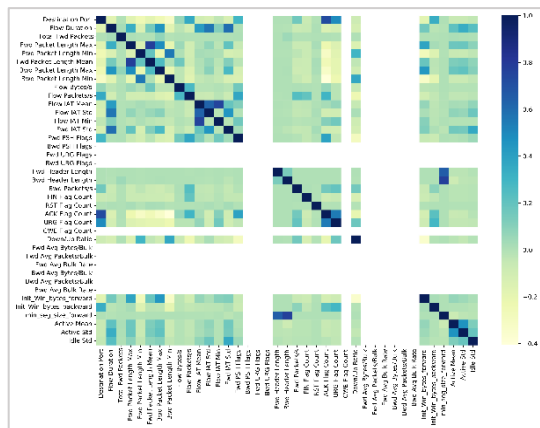

Figure 4. Correlation matrix after selected features based on the threshold <= 0.8 on the training dataset

### 5.1.3. Third Approach

In the third case, one of the XAI techniques is used, namely SHAP (SHapley Additive exPlanation). The features importance are computed for each feature based on the kernel SHAP method, and the top 30 features are selected to build SHAP_Model. The complete approach is discussed in the next subsection, 5.2.

## 5.2. Feature Selection based on Shapley Values for Reconstruction Error

As already discussed, that shapley value for each feature provides the true contribution and cause of the anomalous nature for any attack instance. Therefore this subsection elaborates the complete process of selecting the best features based on shapley values for the reconstruction error of the autoencoder.

The kernelExplainer methodis a model agnostic approach, i.e. independent of the structure of any ML and DL model. This method needs to have access to the dataset and prediction function of the actual model for which we required explanation. The actual autoencoder model, in this case, is based on all features of the CICIDS2017 dataset, trained and validated on benign data only in an unsupervised manner (without target class label). The kernelExplainer required a value function and background dataset to be passed as a parameter to build the explanation model. But the issue is how to define a value function that explains the reconstruction error of the autoencoder. We use the approach proposed by Takeishi et al. [14]. The author proposed two methods. We use one of the methods named Value Function by Marginalization. In this method, the background set (sample of the benign dataset) and mean squared error is passed in the kernelExplainer method to build the explanation model. The value function of the explanation model is defined as in (13) and (14).

$$V(S) = \frac{1}{d} E_{p(x_{S^c}|x_S)}[e(x)] \quad (13)$$

$$x = \begin{bmatrix} x_{S^c} \\ x_S \end{bmatrix} \quad (14)$$

Here, $e(x)$ is the autoencoder reconstruction error, MSE in this case on the instance of the test dataset, $x \in Rd$ it is a vertical concatenation of $x_{S^c}$ and $x_S$. $x_S$ is a subvector of $x$ and $x_{S^c}$ is a complement of $x_S$.

This value function $V(S)$ is used to find the contribution of each feature in the autoencoder reconstruction error using the kernel SHAP method on the sampled CICIDS2017 dataset. And thebackground set consists of 200 benign instances to build the local explanation model. Only attack samples are used to compute the feature importance of each feature based on shapley values. This feature importance plot is the true contribution of each feature affecting the reconstruction error as compared to the raw error of each feature as in Fig. 5. And these features would generally consider being the cause for anomalous behaviour of attack instances. Consequently, we selected the top features to build the final SHAP_Model.

Furthermore, the kernel SHAP can explain the single instance as well as the overall model output. As in Fig. 9, it explains the single benign and attack instance with its reconstruction error.

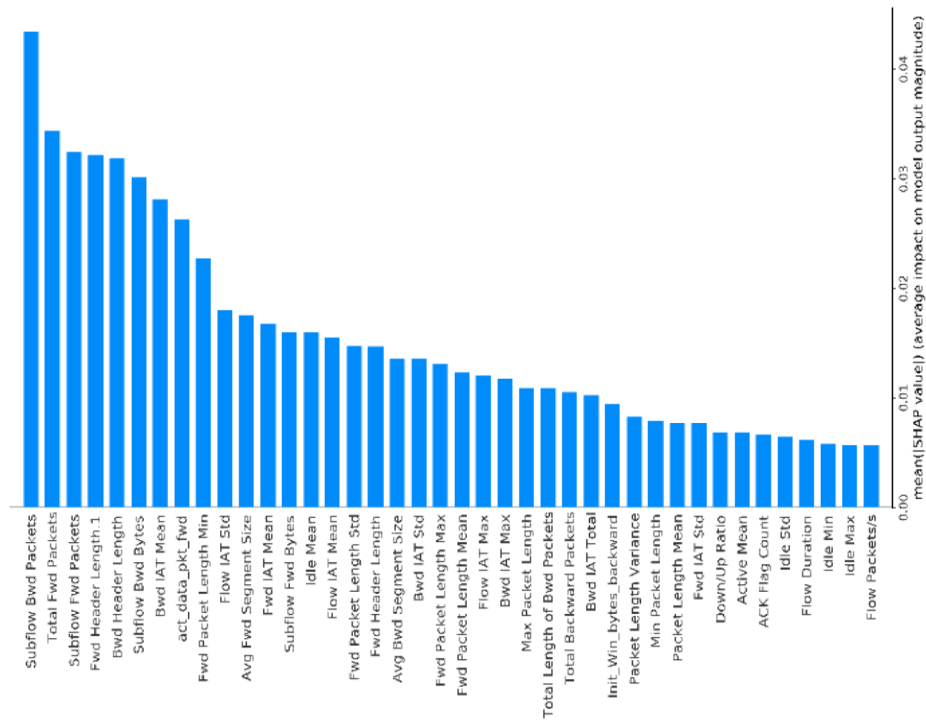

Figure 5. Feature contribution based on shapley values on sample attack dataset

## 5.3. All Models Setup and Evaluation Procedure

All three models are autoencoders artificial neural network architecture (ANN). And the same architecture, such as hidden layers, number of neurons, learning rate, l2 regularizer, is used to compare the results of the three approaches without any bias. However, the initial optimal architecture is selected based on the random search [51]. Random search converges faster than grid search [52] and can perform better with semi optimal set of parameters. The initial architecture is described in Table 3. Learning rates [0.01, 0.001, 0.0001] are selected initially, but the optimal result is achieved on 0.001. In the l2 regularizer parameter [50] same learning rate of 0.001 is passed. ReLU activation function [53] is used in the hidden layers. ReLU is a typical activation function widely used since it removes the vanishing gradient problem of the tanh and sigmoid activation functions [53]. The layered architecture of [70 30 10 30 70] neurons are used and Mean Squared Error (MSE) as loss function.

The complete procedure of building SHAP_Model is illustrated in Fig. 6. Based on the actual autoencoder model (with all features set), the simpler explanation model is created. In the kernelExplainer function, the value function is passed with the sampled background training dataset. Consequently, the built SHAP Explainer is used to plot feature importance graph on sampled attack dataset only. Finally features importance graph is used to select the top 30 features to build SHAP_model on benign data. Fig. 7 illustrates the evaluation procedure for all three models.

Table 3. Autoencoder architecture

| Parameters | AE Architecture |
|---|---|
| Input layer size | size of input features |
| Hidden layer 1 | 70 (neurons) |
| Hidden layer 2 | 30 (neurons) |
| Bottleneck layer 3 | 10(neurons) |
| Hidden layer 4 | 30(neurons) |
| Hidden layer 5 | 70(neurons) |
| Output layer size | size of input features |
| Activation function | relu |
| Optimizer | adam |
| Loss | Mean Squarred Eror (MSE) |
| Learning rate | 1.00E-03 |
| Epochs | 100 |
| Batch_size | 8192 |
| Metrics | accuracy |
| Dataset Split | 67% - 33% |

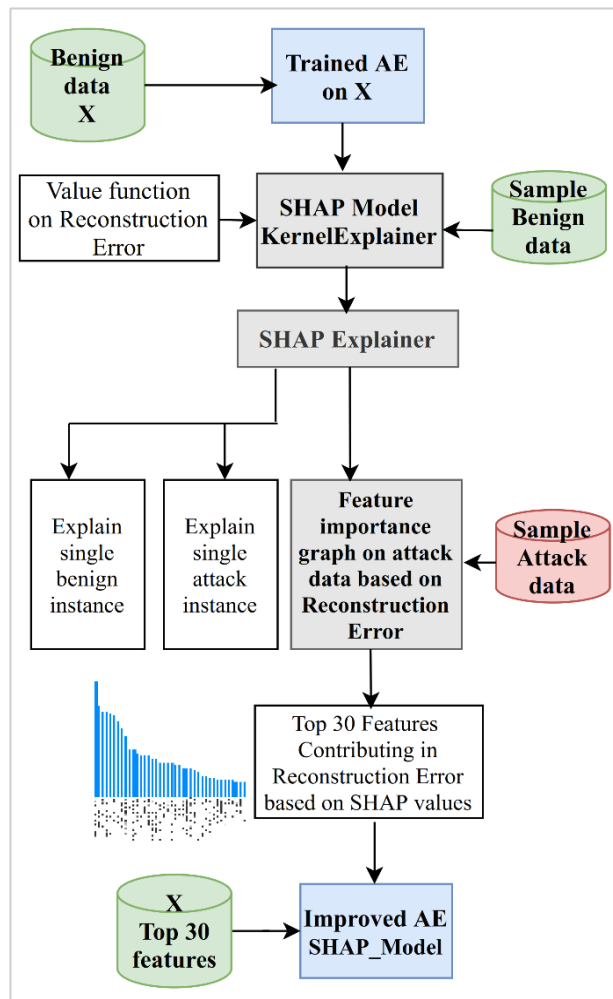

Figure 6. A block diagram of building the SHAP_Model

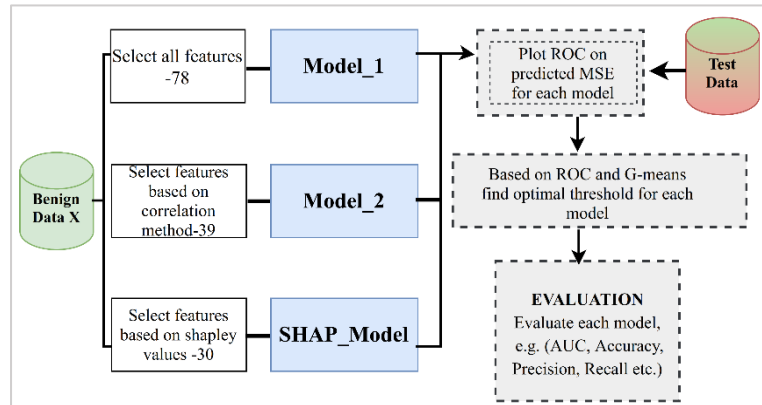

Figure 7. Models building and evaluation procedure

The approach is the same for Model_1, Model_2 and SHAP_Model. The detailed results for ROC, AUR, G-means and Classification report are explained in the Experiment Results section.

## 6. EXPERIMENT RESULTS AND DISCUSSION

This section describes the result of all three AE based models proposed in this paper. As already discussed that Model_1 is based on all features of the CICIDS2017 dataset. Model_2 is based on 39 features selected after the unsupervised correlation approach. And finally, SHAP_Model is based on the top 30 features selected based on shapley values. These top 30 features are highly contributing to the reconstruction error of the model output, causing the anomalous behaviour of the attack instances. Table 4 shows the results of the proposed models.

### 6.1. Results Explaining Output of Autoencoder Reconstruction Error

Fig. 8(a) and 8(b) explain the single normal instance, and 8(c) explain the single attack instance. Here, $E[f(x)] = 0.025$ is base value of the SHAP explanation model and the value is fetched with explainer.expected_value. The base value of the SHAP explainer is the output value of the explainer when all input features have null values. And $f(x)$ is the reconstruction error of the autoencoder. As shown in Fig.8(a) and 8(b) $f(x) = 0.01$ is the reconstruction error of the normal single instance, and in 8(c) $f(x) = 0.73$ is the reconstruction error of the attack instance. Fig. 8 shows that which input feature pushed the model output from the base value of 0.025 to 0.73. Input features that push the output higher are in red, and features pushing it down are in blue.

Shapley values show the true contribution of each feature in the model output. SHAP framework provides many plots such as summary_plot, force_plot, dependency_plot, beeswarm_plots, and much more [54] to visualize and interpret the effect of the model prediction based on input features. Fig. 8(a) is summary plots, and 8(b) & 8(c) are the waterfall_plot of the single normal and attack instance with reconstruction error 0.01 and 0.731, respectively. The feature importance plot for overall model output on the sample background dataset is already described in Section 5. And finally, these features are used to build SHAP_Model to increase its overall results compared to the other two models (Model_1 and Model_2).

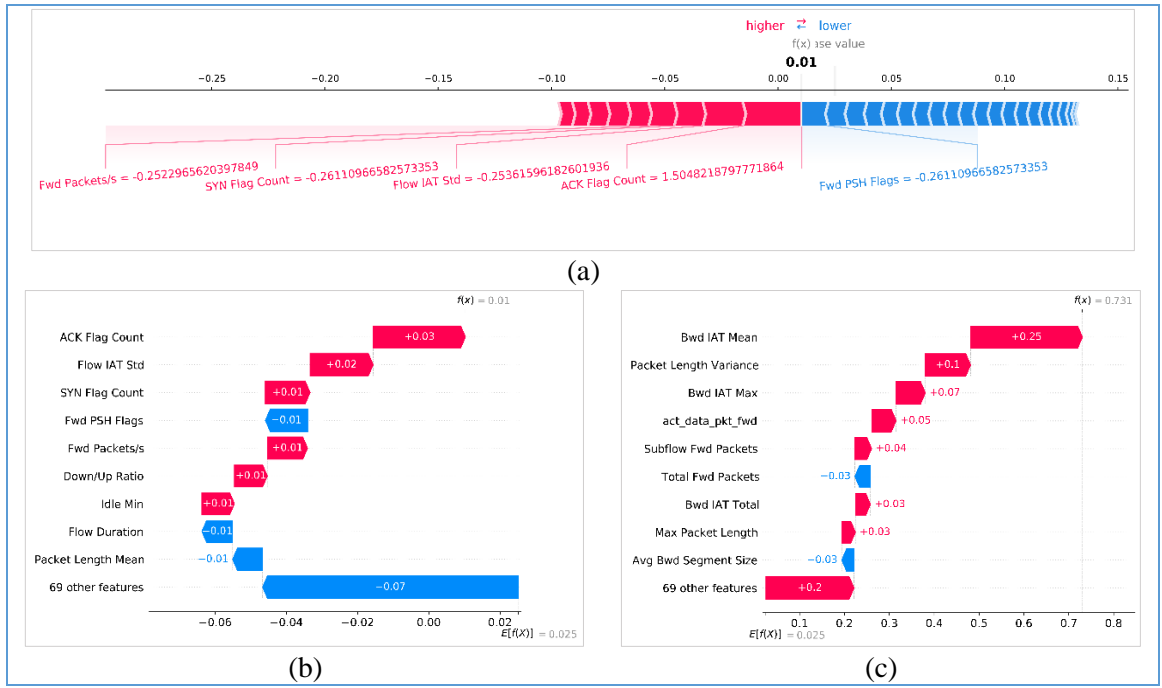

Figure 8. Explaining an normal instance with reconstruction error =0.01 (a) and (b), Explaining an attack instance with reconstruction error =0.731 (c) , E[f(x)] = 0.025 is base value of explaner model (when each feature has null value)

## 6.2. Models Results

All three models are trained and validated on 150000 benign data instances in an unsupervised manner, i.e. the last column corresponds to the class label is not used. And testing data contains a total of 137718 instances, out of which 40000 are anomalies. The architecture of the autoencoder is kept the same for all three models to compare the results without any bias. The optimal batch size is 8192, with a learning rate of 0.001. The difference is only in the feature selection method for the three proposed models. Model_1 is built on all the 78 features, Model_2 is on 39 features based on the unsupervised correlation method, and SHAP_Model is on the top 30 features based on shapley values discussed in Section 5. The extensive experiment is carried out on various metrics, as illustrated in Table 4. The proposed improved SHAP_Model gives great results as compared to the other two models.

The Mean Squared Error is computed on the testing dataset for all three models. Then ROC and AUC (area under the ROC) metrics are used to evaluate performance as they are a good measure for continuous data and unsupervised evaluation [55]. They are also insensitive to the class distribution of the dataset and provide the results on the thresholds that vary from 0.0 to 1.0, as shown in Fig. 9. Furthermore, they are used to demonstrate how one model is better than another on a different threshold. In our method, the ROC is used in selecting the optimal **thresholds [M1-0.17, M2-0.13, SM-0.33]** to compute the classification report for the proposed three models. The G-mean, which is the square root of recall and specificity, is calculated **[M1- 0.812, M2-0.867, SM-0.958]** on ROC to find the best threshold [39]. Then based on the threshold, predicted MSE for each model are converted into binary labels such as [0 = normal, 1= anomaly]. For example, the optimal threshold for Model_1 is 0.17, then the computed MSE less than or equal to 0.17 is labelled as 0 else 1. Finally, the classification report, i.e. Precision, Recall, F-Score and Accuracy, are computed; as a result, SHAP_Model is outperformed compared to other models, as shown in Fig. 10.

International Journal of Computer Networks & Communications (IJCNC) Vol.13, No.6, November 2021

Table 4. All models results

| Model Features Count | G-means | Best Threshold | Precision | Recall | F-score | Accuracy | AUC |
|---|---|---|---|---|---|---|---|
| M1-78 | 0.812 | 0.17 | 0.55 | 0.99 | 0.71 | 0.76 | 0.819 |
| M2-39 | 0.867 | 0.13 | 0.62 | 1 | 0.77 | 0.82 | 0.843 |
| **SM-30** | **0.958** | **0.33** | **0.83** | **0.99** | **0.90** | **0.94** | **0.969** |
| **M1** – Model_1, **M2**—Model_2, **SM**—SHAP_Model ||||||||

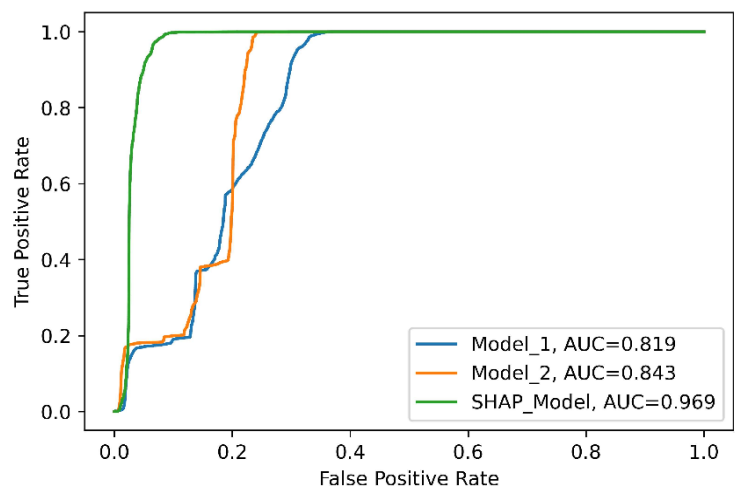

Figure 9. The ROC curve of proposed models on normal and attack data

These results indicate that XAI techniques are so powerful to explain and interpreting the results of complex models such as DL-based models. If we understand the black-box nature of the model, we can further improve the results up to some extent, as we did in this paper. Based on the shapley values the true contribution is computed that further used as feature selection approach to improve the overall result of **SHAP_Model [Precision = 0.83, Recall = 0.99, F-score = 0.90, Accuracy = 0.94, AUC = 0.969].** And the confusion matrix and classification report as in Fig.11 and Fig.12, respectively, shows that out of 40000 anomalies, only 441, i.e. 0.32% out of all the samples, has been misclassified. As the model is trained on only benign data without target class labels, it is a good recall and accuracy score.

### 6.3. Limitations of SHAP Approach

The only drawback we found is that the kernel SHAP is slower and requires a background set (a sample of the dataset). And hence, as we increase the sample size for the background set, the time complexity of the explainer will further increase. In the case of a very high dimensional dataset,the kernel SHAP approach would not be a good choice. Datasets with a large number of features will further increase the computational time to calculate shapley value for each feature and may take several hours in computation. Choosing the correct background set is also important in the kernel SHAP method. The authors of [22] are addressing this issue on their another study.

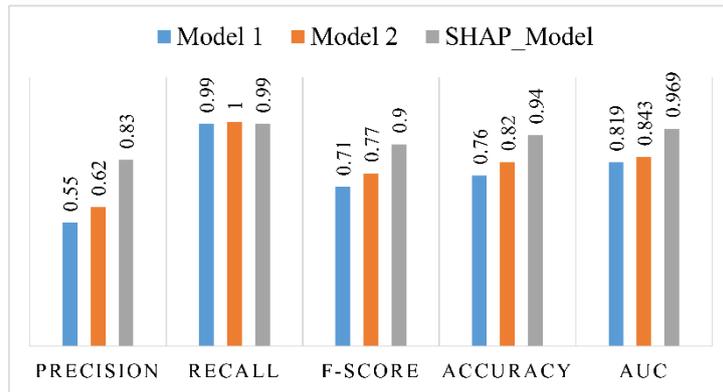

Figure 10. Performance comparison of proposed models

## 6.4. Discussion

In this proposed work, a novel approach of feature selection is presented in an unsupervised manner, i.e. without a target class label. To the best of our knowledge, most of the literature for feature selection is based on supervised and semi-supervised approaches such as Filter methods, Wrapper methods, Embedded methods etc. [56][57]. The kernel SHAP method is used to select the top contributing features in the reconstruction error of the autoencoder-based model to build an optimized model. To further validate and examine our claim, we build three modelsthat clearly indicate that the SHAP_Model significantly improves performance metrics, as shown in Fig. 10. This improvement is due to removing redundant and irrelevant features and choosing only those responsible for anomalous behaviour but without a target class label. The motivation for choosing this approach is due to the lack of availability of labelled datasets in many real-world applications [56] and the emerging XAI approach that further helps in understanding the internal working of black-box models and explaining their results.

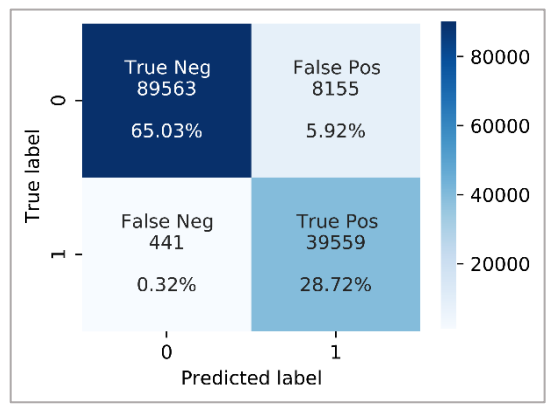

Figure 11. Confusion matrix of SHAP_Model

```
              precision    recall  f1-score   support

         0.0       1.00      0.92      0.95     97718
         1.0       0.83      0.99      0.90     40000

    accuracy                           0.94    137718
   macro avg       0.91      0.95      0.93    137718
weighted avg       0.95      0.94      0.94    137718
```

Figure 12. Classification report of SHAP_Model

## 7. CONCLUSIONS AND FUTURE WORK

This paper proposed SHAP_Model based on the autoencoder for network anomaly detection using shapley values. A subset of the CICIDS2017 dataset is used to do this experiment. The kernel SHAP method, a model agnostics approach, is used to select only those features that are increasing the autoencoder reconstruction error on the attack dataset. The identification of these features is done by computing the shapley values via kennelExplainer (Explanation model) for each feature based on the predicted reconstruction error of the autoencoder model. Shapley values provide the true contribution ofeach feature, causing a large reconstruction error instead of the raw error for each feature. And finally, the top 30 features having large shapley values are used to build SHAP_Model on benign data. These top features are more important than the other features in causing abnormal behaviour of any anomalous instance. SHAP_Models outperformed the other two proposed models, i.e. Model_1 based on all features and Model_2 based on 39 features selected with the unsupervised correlation method.

This work shows that using the available XAI techniques, we can not only explain or interpret the results but can further utilize these techniques to improve the overall performance of any ML/DL model. In our case, we improved the overall performance of the autoencoder based model for network anomaly detection using SHAP. The drawback we found of the kernel SHAP method is its time complexity on the background set. As we increase the background set in the kenelExplainer function, the explanation model will take a lot of time in the building process. Also, choosing the appropriate background set is important in creating an explanation model.

To the best of our knowledge, this paper is the true contribution in essence. And as a future extension, the same method can be tested on other benchmarks and the latest datasets for network anomaly detection. The same approach can be further used in other unsupervised DL architecture such as DBN, RBM etc., to explain and improve their results in network anomaly detection.

### CONFLICT OF INTERESTS

The authors declare no conflict of interest.

International Journal of Computer Networks & Communications (IJCNC) Vol.13, No.6, November 2021

**AUTHORS**

**Ms Khushnaseeb Roshan** completed her B.Sc(Hons) in Computer Science & Application and MCA from the Department of Computer Science, Aligarh Muslim University Aligarh, India. She was the topper student in her Graduation and among the top five students in her Master. She has more than three years of experience in Tata Consultancy Services. She has been pursuing her research in Machine Learning, Deep Learning and Computer Network Security since 2019. Till now, she has presented three Scopus indexed conference papers. She was also awarded JRF(Junior Research Fellowship) at the national level.

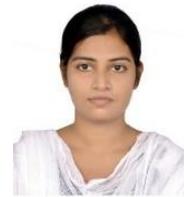

**Dr. Aasim Zafar** is Professor at Computer Science Department, Aligarh Muslim University, Aligarh, India. He holds a Master in Computer Science and Applications and obtained the degree of Ph. D. in Computer Science from Aligarh Muslim University, Aligarh, India. His research areas and special interests include Mobile Ad hoc and Sensor Networks, Software Engineering, Information Retrieval, E-Systems, e-Security, Virtual Learning Environment, Neuro-Fuzzy and Soft Computing, Knowledge Management Systems and Web Mining. His areas of teaching interest include Computer Networks, Network Security, Software Engineering, E-Systems, Database Management Systems and Computer Programming. He has presented many papers in National and International Conferences and published various research papers in journals of international repute.

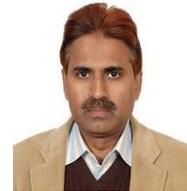